\documentclass[11pt]{article}

\usepackage[final]{acl}

\usepackage{times}
\usepackage{latexsym}

\usepackage[T1]{fontenc}
\usepackage[utf8]{inputenc}

\usepackage{microtype}

\usepackage{inconsolata}

\usepackage{graphicx}

\usepackage{amsmath,amssymb,amsfonts}
\usepackage{algorithmic}
\usepackage{graphicx}
\usepackage{textcomp}
\usepackage{xcolor}
\usepackage{booktabs}
\usepackage{multirow}
\usepackage{makecell}
\usepackage{colortbl}
\usepackage{pifont}
\usepackage{xspace}

\newcommand{\modelname}{Libra-VLA\xspace}
\newcommand{\sysone}{Semantic Planner\xspace}
\newcommand{\systwo}{Action Refiner\xspace}

\title{Libra-VLA: Achieving Learning Equilibrium via Asynchronous Coarse-to-Fine Dual-System}

\author{
  \textbf{Yifei Wei\textsuperscript{1,2}},
  \textbf{Linqing Zhong\textsuperscript{1,2}},
  \textbf{Yi Liu\textsuperscript{2}},
  \textbf{Yuxiang Lu\textsuperscript{2}},
  \textbf{Xindong He\textsuperscript{2}},
\\
  \textbf{Maoqing Yao\textsuperscript{2}\thanks{Corresponding authors.}},
  \textbf{Guanghui Ren\textsuperscript{2}\footnotemark[1]}
\\
\\
  \textsuperscript{1}Beihang University,
  \textsuperscript{2}AgiBot
\\
}

\begin{document}
\maketitle
\begin{abstract}
Vision-Language-Action (VLA) models are a promising paradigm for generalist robotic manipulation by grounding high-level semantic instructions into executable physical actions. However, prevailing approaches typically adopt a \textit{monolithic generation paradigm}, directly mapping visual-linguistic features to high-frequency motor commands in a flat, non-hierarchical fashion. This strategy overlooks the inherent hierarchy of robotic manipulation, where complex actions can be naturally modeled in a Hybrid Action Space, decomposing into \textit{discrete} macro-directional reaching and \textit{continuous} micro-pose alignment, severely widening the semantic-actuation gap and imposing a heavy representational burden on grounding high-level semantics to continuous actions.
To address this, we introduce \modelname, a novel Coarse-to-Fine Dual-System VLA architecture. We explicitly decouple the learning complexity into a coarse-to-fine hierarchy to strike a training equilibrium, while simultaneously leveraging this structural modularity to implement an asynchronous execution strategy.
The \sysone predicts discrete action tokens capturing macro-directional intent, while the \systwo conditions on coarse intent to generate high-frequency continuous actions for precise alignment.
Crucially, our empirical analysis reveals that performance follows an inverted-U curve relative to action decomposition granularity, peaking exactly when the learning difficulty is balanced between the two sub-systems. With the asynchronous design, our approach offers a scalable, robust, and responsive solution for open-world manipulation. Project page: \url{https://libra-vla.github.io/}.
\end{abstract}

\section{Introduction}

\begin{figure}[t]
    \centering
    \includegraphics[width=\columnwidth]{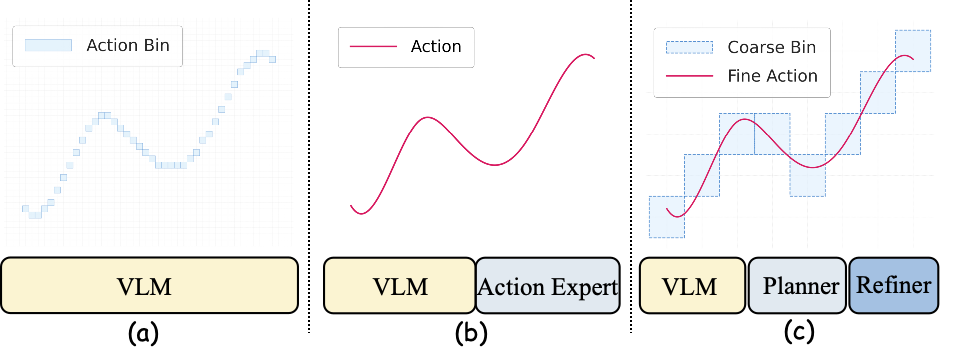}
    % \vspace{-5.5mm}
    \caption{Comparison of action generation paradigms. (a) Discrete autoregressive approaches discretize actions into massive bins. (b) Continuous diffusion approaches directly predict continuous signals. (c) Our proposed Libra-VLA operates in a hybrid action space, where discrete coarse bins representing macro-intents serve as anchors for continuous fine actions, naturally aligning with the inherent hierarchical characteristics.}
    \vspace{-6mm}
    \label{fig:paradigm_comparison}
\end{figure}

The pursuit of generalist robots capable of performing diverse manipulation tasks in open-world environments remains a central challenge in embodied intelligence~\cite{zhao2023learning}. Recent advancements in Large Vision-Language Models (VLMs), trained on internet-scale corpora, have endowed systems with unprecedented capabilities in visual understanding and semantic reasoning~\cite{bai2025qwen2}. Through extending these pre-trained backbones into the robotic domain, Vision-Language-Action (VLA) models have emerged as a dominant paradigm~\cite{gr00t}.
Unlike traditional specialist policies limited to narrow tasks, VLAs demonstrate remarkable potential in grounding abstract natural language instructions into physical actions, enabling robots to generalize across novel objects and scenarios~\cite{intelligence2025pi_}. 
However, bridging the gap between semantic intelligence and physical actuation requires more than a direct alignment. Naturally, physical manipulation is not a singular, atomic event, but a process with inherent structure—typically progressing from broad, semantic-driven approaches to precise, geometry-driven interactions.
Despite this physical reality, the prevailing VLA paradigm typically employs a \textit{monolithic generation strategy} that overlooks this hierarchical nature. Specifically, previous approaches often adopt direct mapping from high-level semantic features to low-level motor commands through a unified architecture, as shown in Fig.~\ref{fig:paradigm_comparison} (a) and (b). 

These methods typically operate by either discretizing continuous actions into numerous action bins to approximate the continuous action space~\cite{kim2024openvla}, or by attaching continuous diffusion heads to the VLM backbone to directly predict continuous signals~\cite{black2410pi0}. Such enforced uniformity creates a massive semantic-actuation gap~\cite{zhang2025vlaseffectivelyinheritvlms,intelligence2025pi_}. The model is compelled to simultaneously grapple with high-level abstract reasoning and low-level high-frequency control within a flat, non-hierarchical process. A natural remedy is to introduce hierarchical control. Existing approaches primarily focus on \textit{temporal decomposition}, guiding task execution through sequential milestones such as sub-instructions or goal images~\cite{hamster,hirobot}. However, temporal decoupling only shortens the planning horizon. The model still performs a direct cross-modal translation to continuous actions, leaving the representational complexity of single-step generation unresolved. 

In contrast to existing temporal decompositions, we advocate for a paradigm shift: extending the hierarchical philosophy to the representational action space. We propose modeling robotic action within a \textbf{Hybrid Action Space} (Fig.~\ref{fig:paradigm_comparison} (c)), decomposing complex behaviors into discrete macro-directional reaching for semantic grounding and continuous micro-pose alignment for geometric precision.
Rather than striving to minimize quantization errors by indiscriminately increasing the number of action bins, we strategically leverage the inherent coarseness of this discrete space to represent abstract \textit{macro-directional intents}.
Intuitively, this hybrid decomposition acts as a progressive bridge across the modality gap: the discrete macro-intent resolves the high-level semantic ambiguity of \textit{``where to go''}, providing a stable geometric anchor that constrains the search space. Conditioned on this anchor, the continuous micro-alignment is liberated to focus exclusively on \textit{``how to interact''}, synthesizing the high-frequency residuals necessary for interaction-rich tasks.
Ultimately, this formulation allows the discrete subspace to align with high-level linguistic semantics, while the continuous subspace ensures the fidelity required for complex physical execution.

Based on this insight, we introduce \textbf{Libra-VLA}, which instantiates the coarse-to-fine philosophy through a decoupled dual-system architecture. Instead of compelling a monolithic network to bridge the extensive semantic-actuation gap, our proposed framework distributes the computational complexity across two specialized modules: \sysone and \systwo.
Technically, \sysone focuses on the discrete subspace, augmenting a general-purpose VLM with a parallel decoding transformer to predict coarse directional tokens. This design leverages the backbone's spatial reasoning to resolve semantic ambiguity without being burdened by metric precision. Conversely, the \systwo handles the continuous subspace. It employs a diffusion transformer equipped with an independent high-resolution visual encoder to capture local geometric details, synthesizing fine-grained residuals conditioned on the planner's intent. Beyond training stability, this structural modularity naturally facilitates an asynchronous execution strategy: \sysone operates at a lower frequency to provide stable guidance, while \systwo executes at high frequency, ensuring real-time responsiveness.

To summarize, our main contributions are as follows:

\begin{itemize}

    \item We propose a novel Coarse-to-Fine VLA paradigm grounded in a hybrid action space and identify the principle of learning complexity equipartition, demonstrating that model performance follows an inverted-U curve and peaks exactly when the learning difficulty is balanced between the two phases.

    \item Building on this hybrid paradigm, we implement a decoupled asynchronous dual-system architecture. This design allows the \sysone to operate at a low frequency for stable discrete planning, while the \systwo executes at a high frequency for real-time continuous control.

    \item We develop \modelname and demonstrate that it outperforms baselines by achieving higher success rates and lower inference latency.

\end{itemize}

\section{Related works}
\subsection{Hierarchical Generation}

Hierarchical control has a long-standing tradition in robotics. Option learning~\cite{stolle2002options} discovers temporally extended sub-policies by identifying useful subgoals, while hierarchical reinforcement learning~\cite{2019hrl} leverages language as the abstraction to decompose long-horizon tasks. Recent hierarchical VLA models inherit this temporal decomposition philosophy: HAMSTER~\cite{hamster} and MOKA~\cite{MOKA} predict keypoints or waypoints as intermediate sub-goals, while ViLA~\cite{ViLA} and Hi Robot~\cite{hirobot} generate step-by-step language sub-instructions to guide low-level policies. While effective for long-horizon planning, these approaches universally operate along the temporal axis. A common limitation is that the inter-level communication resides in a different modality from the final motor commands, forcing the low-level policy to perform a cross-modal translation that introduces a severe modality gap.

Our work circumvents this bottleneck by introducing the coarse action as an intra-modal intermediate state that progressively bridges the modality gap through two simplified mappings. First, mapping VLM features to coarse actions with a small vocabulary size is formulated as a low-cardinality discrete classification task, which substantially reduces the alignment difficulty. Second, mapping coarse actions to fine actions operates entirely within the same modality, where the coarse intent serves as a geometric anchor that drastically narrows the search space for final continuous action generation. We note that HybridVLA~\cite{liu2025hybridvlacollaborativediffusionautoregression} also models actions in a hybrid space, yet both of its branches independently predict fine-grained actions and are fused via arithmetic averaging, constituting a flat parallel architecture without hierarchical structure.

\subsection{Dual-System VLA Architectures}

Inspired by cognitive dual-process theory~\cite{kahneman2011thinking, evans2008dual}, several recent works~\cite{cui2025openhelixshortsurveyempirical,gr00t,fis,lcb} adopt a Dual-System architecture that decouples manipulation into a slow, deliberate System 2 for high-level reasoning and a fast, intuitive System 1 for low-level execution. However, current implementations exhibit shared structural bottlenecks. Models such as GR00T N1~\cite{gr00t} rely on static latent embeddings as the inter-system communication bridge. During asynchronous generation, these features lack future temporal context and become increasingly lagging as the environmental states change. In contrast, our planner generates a predictive sequence of coarse actions to cover the upcoming execution horizon. This enables effective asynchronous execution via a predictive intent buffer, providing temporally synchronized guidance for each time step. Meanwhile, architectures like FiS-VLA~\cite{fis} force feature coupling within a single backbone, imposing a heavy representational burden by forcing shared network weights to simultaneously encode high-level semantics and low-level fine-grained features. We instead equip the fast system with an independent visual encoder, achieving structural decoupling that eliminates this feature-squeezing bottleneck. Moreover, existing dual systems such as OpenHelix~\cite{cui2025openhelixshortsurveyempirical} universally employ high-dimensional black-box latent vectors for inter-system communication, rendering the information flow between subsystems opaque and difficult to interpret. Our framework replaces these implicit latents with explicit, directly executable coarse actions that carry clear physical semantics as macro-directional intents, yielding a transparent and interpretable communication protocol between the planner and the refiner.

\section{Methodology}
\subsection{Problem Formulation}
\label{sec:problem_formulation}

\begin{figure*}[t]
  \includegraphics[width=\linewidth]{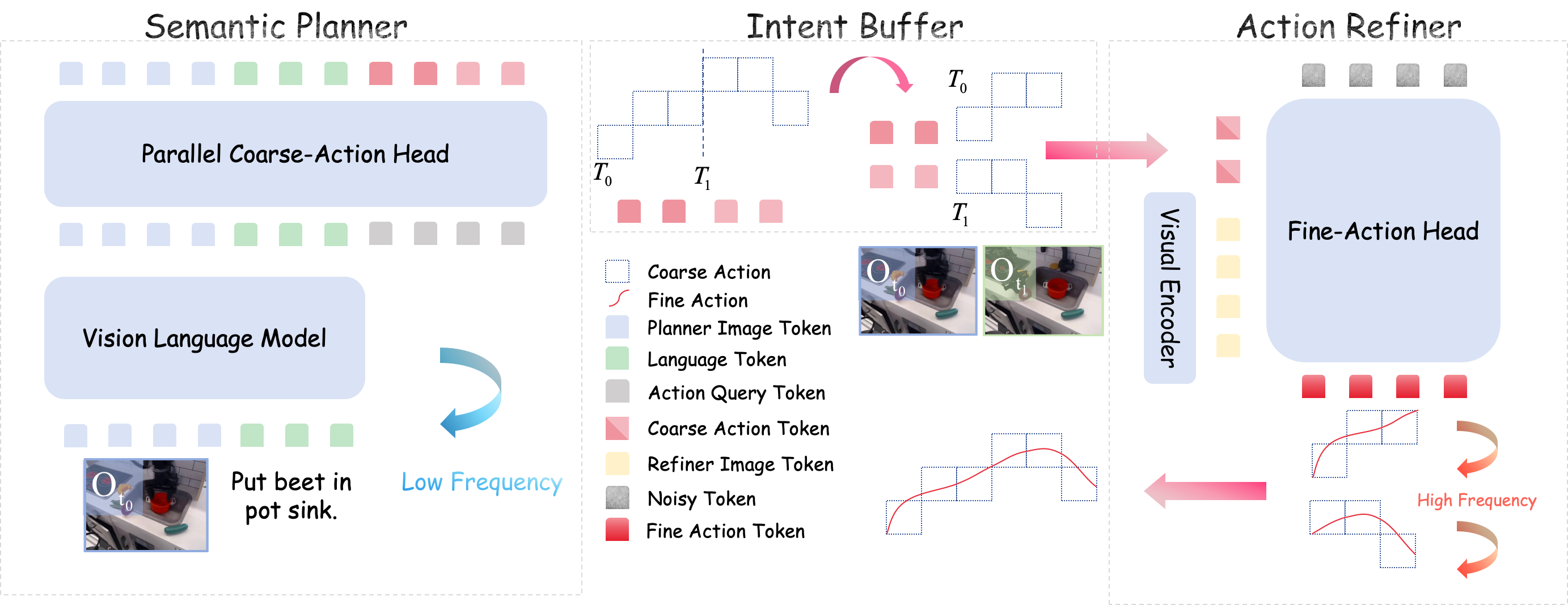}
  \caption{
  Architectural Overview of Libra-VLA.
  The framework adopts a Coarse-to-Fine generation paradigm, explicitly modeling actions within a \textbf{Hybrid Action Space} across two specialized phases.
  (Left) System 2: \sysone runs at a low frequency, employing a VLM backbone augmented with a Parallel Coarse-Action Head to predict discrete macro-directional intents (coarse action tokens) within the discrete semantic subspace. 
  (Right) System 1: \systwo runs at a high frequency, utilizing a diffusion transformer equipped with an independent visual encoder to synthesize continuous micro-pose alignments (fine action tokens) within the continuous geometric subspace. 
Crucially, the two systems are bridged via an asynchronous execution strategy managed by an intent buffer. Controlled by the Horizon Expansion Factor ($M$), the Semantic Planner anticipates an extended coarse action chunk in a single inference pass. Subsequently, the Action Refiner iteratively retrieves the specific coarse action slice corresponding to the current timestep from the buffer as the condition for its high-frequency generation.
}
  \label{fig:fig1}
  \vspace{-5mm}
\end{figure*}

We formulate the VLA problem as learning a policy $\pi$ that maps instructions $L$ and observations $\mathbf{o}_t$ to continuous actions $\mathbf{a}_t \in \mathcal{A}$. Specifically, we decompose the action generation into a discrete coarse-grained intention and a continuous fine-grained action:

\begin{equation}
    \mathbf{a}_t = \Phi(\mathbf{a}_t^{c}, \mathbf{a}_t^{f}),
\end{equation}

\noindent where:
\begin{itemize}
    \item $\mathbf{a}_t^{c} \in \mathcal{V}_{act}$ resides in the discrete semantic subspace, capturing the broad directional intent to align with VLM reasoning.
    \item $\mathbf{a}_t^{f} \in \mathbb{R}^{d}$ resides in the continuous geometric subspace, responsible for precise geometric adjustments.
\end{itemize}

Based on this decomposition, we factorize the joint policy distribution using the probabilistic chain rule. The probability of generating the final action $\mathbf{a}_t$ is decoupled into two conditional distributions:

\begin{equation}
    \label{eq:factorization}
    P(\mathbf{a}_t | \mathbf{o}_t, L) \approx \underbrace{P(\mathbf{a}_t^{f} | \mathbf{a}_t^{c}, \mathbf{o}_t)}_{\text{Action Refiner}} \cdot \underbrace{P(\mathbf{a}_t^{c} | \mathbf{o}_t, L).}_{\text{Semantic Planner}}
\end{equation}

This factorization formally underpins our dual-system architecture: System 2 (\sysone) is optimized to maximize the likelihood of the coarse intent $P(\mathbf{a}_t^{c} | \mathbf{o}_t, L)$ by leveraging the reasoning capabilities of VLMs, while System 1 (\systwo) models the conditional distribution $P(\mathbf{a}_t^{f} | \mathbf{a}_t^{c}, \mathbf{o}_t)$ via a generative diffusion process to synthesize precise actions. The overall framework of Libra-VLA is shown in Fig.~\ref{fig:fig1}.

\subsection{System 2: \sysone}
\label{sec:systeLibra-Base}
\noindent\textbf{Coarse-Grained Directional Discretization.}
To cast physical actions into a format natively compatible with the VLM's discrete output space, we discretize the normalized continuous actions into $N$ uniform bins. Specifically, let $\mathbf{a}_t \in [-1,1]^{D}$ denote the normalized continuous action vector at timestep $t$, where $D$ is the dimensionality of the action space. The ground truth discrete index for the $i$-th dimension, $y_{t,i}^{gt} \in \{0,1,\dots,N-1\}$, is obtained via uniform quantization:
\begin{equation}
    y_{t,i}^{gt} = \operatorname{clip}\left(\left\lfloor \frac{a_{t,i}+1}{2} \times N \right\rfloor,\; 0,\; N-1\right),
\end{equation}
where $(a_{t,i}+1)/2$ maps the action value from $[-1,1]$ to $[0,1]$, the multiplication by $N$ scales it to the bin range, $\lfloor\cdot\rfloor$ assigns the integer bin index, and $\operatorname{clip}$ handles boundary conditions.
Distinct from previous VLA works~\cite{kim2024openvla} that typically utilize fine-grained binning ($N=256$) to approximate continuous control with high precision, we deliberately employ a significantly smaller number of bins ($N \ll 256$). Rather than pursuing precise kinematics, this coarse discretization abstracts actions into macro-directional intents that naturally align with the VLM's semantic reasoning capabilities. Meanwhile, the significantly reduced number of bins narrows the discrete action vocabulary, effectively alleviating the learning burden on \sysone. A detailed quantitative analysis is presented in the ablation studies.

\noindent\textbf{Parallel Coarse-Action Head.}
The primary role of \sysone is to generate a \textit{coarse-grained directional intent} $\mathbf{a}_t^{c}$ via a \textbf{Parallel Coarse-Action Head} to accelerate inference. We introduce learnable query tokens $\mathbf{Q}_{act} \in \mathbb{R}^{K \times D}$, which are concatenated with the VLM features $\mathbf{H}_t$ and fed into a bidirectional transformer. The output tokens corresponding to the query positions are sliced to obtain the refined features $\mathbf{Z}_{act}$, which are then projected to probability distributions over the $N$ discrete bins:

\begin{equation}
\begin{split}
    \mathbf{Z}_{act} &= \operatorname{Self-Attention}([\mathbf{Q}_{act}; \mathbf{H}_t])_{0:K}, \\
    P(\mathbf{a}_t^{c}) &= \operatorname{Softmax}\left( \operatorname{Linear}\left( \mathbf{Z}_{act} \right) \right).
\end{split}
\end{equation}

The training objective of \sysone is to minimize the standard Cross-Entropy loss between the predicted probability distribution $P(\mathbf{a}_t^{c})$ and the ground truth discrete action indices $\mathbf{y}_t^{gt}$ derived from the quantization strategy. Formally, the loss function is defined as:
\begin{equation}
    \mathcal{L}_{plan} = \mathcal{L}_{CE}\left( P(\mathbf{a}_t^{c}), \mathbf{y}_t^{gt} \right).
\end{equation}

To summarize, \sysone yields a probability distribution over discrete bins representing the macro-directional intent, which subsequently serves as the conditional geometric anchor for the fine-grained action synthesis in the next stage.

\subsection{System 1: \systwo}
\label{sec:systeLibra-VE}

While System 2 provides the high-level semantic roadmap, the execution of manipulation tasks requires continuous and precise motor commands. To achieve this, System 1 operates as a conditional diffusion policy that refines the coarse intent into executable precise motions.

\noindent\textbf{Adaptive Intent Injection.}
The core function of System 1 is to synthesize fine-grained actions conditioned on macro-intent provided by System 2. To instantiate this embedding, we maintain a learnable codebook $\mathbf{E} \in \mathbb{R}^{N \times D}$, where $N$ corresponds to the bin size defined in the quantization strategy.
Distinct from standard teacher-forcing methods, we implement a dynamic curriculum strategy for determining the source of $\mathbf{e}_{intent}$ during training, which evolves based on the performance of System 2.

System 1 synthesizes fine-grained actions conditioned on the macro-intent embedding $\mathbf{e}_{intent}$ retrieved from the codebook $\mathbf{E}$. To bridge the training-inference gap, we implement a dynamic curriculum strategy. In early stages (when the prediction success rate of System 2 is below a given threshold $\tau$), $\mathbf{e}_{intent}$ is retrieved using the ground-truth discrete tokens to stabilize training. As the planner improves, we switch to obtaining $\mathbf{e}_{intent}$ by sampling from the predicted distribution $P(\mathbf{a}_t^{c})$. This strategy exposes the refiner to planning noise, effectively fostering an inherent error-correction capability against minor directional deviations.

\noindent\textbf{Precise Action Generation.}
We model the precise action generation as a conditional denoising process. The core architecture of \systwo is implemented as a diffusion transformer, fully utilizing its ability to model complex multi-modal distributions. To furnish sufficiently fine-grained visual representations for precise actuation while achieving structural decoupling from \sysone, we augment the Fine-Action Head with an auxiliary visual encoder $\mathcal{E}_{vis}$ to extract geometric features $\mathbf{F}_t^{geo} = \mathcal{E}_{vis}(\mathbf{o}_t)$. Subsequently, the Fine-Action Head conditions on the composite input of the noisy action $\mathbf{x}_k$, the geometric features $\mathbf{F}_t^{geo}$, and the macro-intent $\mathbf{e}_{intent}$ to predict the noise $\epsilon_\theta(\mathbf{x}_k, \mathbf{F}_t^{geo}, \mathbf{e}_{intent})$, thereby iteratively reversing the diffusion process to recover the robot action $\mathbf{a}_t^f$.

The entire System 1 is trained to minimize the standard Mean Squared Error between the predicted noise and the actual noise added during the forward diffusion process. The loss function is defined as:
\begin{equation}
    \mathcal{L}_{diff} = \mathbb{E}_{k, \mathbf{x}_0, \epsilon} \left[ \| \epsilon - \epsilon_\theta(\mathbf{x}_k, \mathbf{F}_t^{geo}, \mathbf{e}_{intent}) \|^2 \right].
\end{equation}

To jointly optimize the hierarchical architecture, the overall loss function is defined as follows:

\begin{equation}
    \mathcal{L}_{total} = \lambda_{diff} \mathcal{L}_{diff} + \lambda_{plan} \mathcal{L}_{plan},
\end{equation}
where $\lambda_{diff}$ and $\lambda_{plan}$ are calibrated to balance the loss magnitudes, thereby preventing gradient dominance.

\subsection{Asynchronous Execution Strategy}
\label{sec:async_execution}

To mitigate the computational burden of executing the expensive VLM at every inference cycle, we explicitly decouple high-level reasoning from low-level control via an asynchronous execution strategy managed by a \textit{semantic intent buffer}. This mechanism allows System 2 to plan periodically in bursts while System 1 operates continuously at high frequency.

\noindent\textbf{Planning Horizon Expansion.} To enable asynchronous execution, we configure \sysone to predict a longer macro-horizon $L_{macro} = M \times H_{chunk}$ in a single inference pass, where $H_{chunk}$ is the execution horizon of \systwo. Here, $M$ serves as the \textit{Horizon Expansion Factor}, representing the ratio of control to planning frequency. This allows \sysone to encapsulate the directional intent for $M$ subsequent execution chunks, reducing the planning frequency.

\noindent\textbf{Intent Buffering and Consumption.}
We build a First-In-First-Out (FIFO) queue, denoted as the Intent Buffer $\mathcal{Q}$, to bridge \sysone and \systwo. The execution workflow proceeds as follows:

Buffer Refill (Low Frequency): At the beginning of a cycle, if the buffer $\mathcal{Q}$ is empty, System 2 performs a forward pass given the current observation $\mathbf{o}_t$ and instruction $L$. It generates $L_{macro}$ coarse tokens, which are immediately pushed into $\mathcal{Q}$. Crucially, during the subsequent remaining $M-1$ control steps, System 2 remains dormant, bypassing the time-consuming VLM inference.
    
Conditional Consumption (High Frequency): System 1 operates at the robot's control frequency. At each step $k$, instead of querying the VLM, System 1 retrieves the corresponding slice of coarse tokens from the buffer:
    \begin{equation}
        \mathbf{a}_{slice}^c = \mathcal{Q}.\text{pop}(H_{chunk}).
    \end{equation}
This retrieved slice $\mathbf{a}_{slice}^c$ serves as the condition $\mathbf{e}_{intent}$ for precise action generation. System 1 then denoises the fine-grained actions for current timestep.

\begin{table}[t]
    \centering
    \renewcommand{\arraystretch}{1.1} 
    \resizebox{\columnwidth}{!}{
        \begin{tabular}{l|c|c|c|c|c|c}
        \toprule
        Methods & Action Space & Spatial & Object & Goal & Long & Avg. \\
        \midrule
        
        CoT-VLA~\cite{zhao2025cot} & Discrete & 
        87.5 & 91.6 & 87.6 & 69.0 & 81.1 \\
        
        WorldVLA~\cite{cen2025worldvla} & Discrete & 
        87.6 & 96.2 & 83.4 & 60.0 & 81.8 \\
        
        DD-VLA~\cite{liang2025discrete} & Discrete &
        97.2 & 98.6 & \underline{97.4} & 92.0 & 96.3 \\

        OpenVLA~\cite{kim2024openvla} & Discrete &
        84.7 & 88.4 & 79.2 & 53.7 & 76.5 \\

        $\pi_{0}$-FAST~\cite{pertsch2025fast} & Discrete &
        96.4 & 96.8 & 88.6 & 60.2 & 85.5 \\

        Diffusion Policy~\cite{chi2025diffusion} & Continuous &
        78.3 & 92.5 & 68.3 & 50.5 & 72.4 \\

        Octo~\cite{team2024octo} & Continuous &
        78.9 & 85.7 & 84.6 & 51.1 & 75.1 \\

        DreamVLA~\cite{zhang2025dreamvla} & Continuous &
        97.5 & 94.0 & 89.5 & 89.5 & 92.6 \\

        F1~\cite{lv2025f1} & Continuous &
        98.2 & 97.8 & 95.4 & 91.3 & 95.7 \\

        GR00T-N1~\cite{gr00t} & Continuous &
        94.4 & 97.6 & 93.0 & 90.6 & 93.9 \\
        
        GO-1~\cite{bu2025agibot_iros}& Continuous & 
        96.2 & 97.8 & 96.0 & 89.2 & 94.8 \\

        GE-Act~\cite{liao2025genie}& Continuous & 
        98.2 & 97.6 & 95.8 & \textbf{94.4} & 96.5 \\

        $\pi_{0}$~\cite{black2410pi0} & Continuous & 
        96.8 & \underline{98.8} & 95.8 & 85.2 & 94.1 \\

        $\pi_{0.5}$~\cite{intelligence2025pi_} & Continuous & 
        \textbf{98.8} & 98.2 & \textbf{98.0} & 92.4 & \underline{96.9} \\

        \midrule
        \rowcolor{gray!20}
        \textbf{Ours} & \textbf{Hybrid} & 
        \underline{98.6} & 
        \textbf{99.4} & 
        \textbf{98.0} & 
        \underline{92.8} & 
        \textbf{97.2} \\
        \bottomrule
    \end{tabular}
    }
    \caption{Comparison on the LIBERO benchmark. The best results are highlighted in \textbf{bold}, and the second-best results are \underline{underlined}.}
    \vspace{-4mm}
    \label{tab:libero}
\end{table}
\section{Experiments}

In this section, we present a comprehensive empirical evaluation of our proposed \modelname. To rigorously evaluate the model's performance in terms of both precise manipulation and robustness, we utilize two simulation benchmarks: LIBERO~\cite{liu2023libero} for assessing standard capabilities, and LIBERO-Plus~\cite{fei2025libero} for conducting an in-depth analysis. Furthermore, to demonstrate the effectiveness of \modelname in physical world, we also conducted a series of real-world experiments. More details about these simulation benchmarks and real-world tasks are introduced in Appendix~\ref{appendix:training_data}.

\subsection{Experimental Setup}
\noindent\textbf{Model Implementation.} We initialize the VLM backbone of \modelname with InternVL2.5-2B. Notably, all simulation and real-world experiments in this paper are conducted without large-scale robot-data pretraining.
Structurally, both the Parallel Coarse-Action Head and the Fine-Action Head are implemented as transformer blocks comprising $N=12$ attention layers, the hidden state dimension of which is set to $1024$, corresponding to half of the VLM backbone's hidden size. We employ SigLIP~\cite{zhai2023sigmoid} as the visual encoder within the Action Refiner. Unless explicitly stated otherwise, all results on simulation benchmarks are obtained with a Horizon Expansion Factor of $M=2$ and an action chunk size of $H_{chunk}=5$, resulting in a macro-horizon of $L_{macro}=10$. More details about model implementation and training configuration are introduced in the Appendix~\ref{appendix:model_implementation} and Appendix~\ref{appendix:training_details} respectively.

\subsection{Simulation Experiments}
\noindent\textbf{LIBERO Benchmark.}
LIBERO serves as the primary test for evaluating the capabilities of generalist robot policies. It provides a procedural generation pipeline comprising multiple diverse manipulation tasks, categorized into four distinct task suites: LIBERO-Spatial, LIBERO-Object, LIBERO-Goal, and LIBERO-Long. The benchmark is designed to provide a comprehensive, multifaceted evaluation of the model's capabilities, encompassing spatial understanding, object manipulation, instruction following, and long-horizon task execution. Following standard evaluation protocols in prior works, each task is evaluated 50 times independently, 500 rollouts total for the task suites. We report the success rates on the four task suites, as well as the average success rate across all four tasks.

As shown in Table~\ref{tab:libero}, \modelname establishes a new state-of-the-art with an average success rate of 97.2\%, significantly outperforming baselines. Specifically, the framework achieves dominant performance on precision-critical tasks (99.4\% on Object), validating the fine-grained geometric control of our \systwo. It also excels in complex long-horizon tasks (92.8\% on Long), underscoring the robust macro-directional guidance provided by the \sysone.

\begin{table*}[t]
    \centering
    \renewcommand{\arraystretch}{1.1}
    \setlength{\tabcolsep}{2.5pt}
    \resizebox{\textwidth}{!}{%
    \begin{tabular}{l|c|ccccccc|c}
        \toprule
        Methods & Action Space &
        Camera & Robot & Language & Light & Background & Noise & Layout & Avg. \\

        \midrule
        \midrule
        \multicolumn{10}{c}{\textit{Zero-Shot Transfer}} \\
        \midrule
        \midrule

        WorldVLA~\cite{cen2025worldvla} & Discrete &
        0.1 & 27.9 & 41.6 & 43.7 & 17.1 & 10.9 & 38.0 & 25.0 \\
        OpenVLA~\cite{kim2024openvla} & Discrete &
        0.8 & 3.5 & 23.0 & 8.1 & 34.8 & 15.2 & 28.5 & 15.6 \\
        NORA~\cite{hung2025nora} & Discrete &
        2.2 & 37.0 & 65.1 & 45.7 & 58.6 & 12.8 & 62.1 & 39.0 \\
        UniVLA~\cite{bu2025univlalearningacttaskcentric} & Continuous &
        1.8 & \underline{46.2} & 69.6 & 69.0 & 81.0 & 21.2 & 31.9 & 42.9 \\
        $\pi_0$-Fast~\cite{pertsch2025fast} & Discrete &
        \underline{65.1} & 21.6 & 61.0 & 73.2 & 73.2 & 74.4 & 68.8 & 61.6 \\
        OpenVLA-OFT~\cite{kim2025fine} & Continuous &
        56.4 & 31.9 & \underline{79.5} & \underline{88.7} & \underline{93.3} & \underline{75.8} & \underline{74.2} & \underline{69.6} \\

        \midrule
        \rowcolor{gray!20}
        \textbf{Ours} & \textbf{Hybrid} &
        \textbf{68.9} &
        \textbf{48.8} &
        \textbf{92.7} &
        \textbf{97.9} &
        \textbf{93.4} &
        \textbf{86.3} &
        \textbf{77.5} &
        \textbf{79.5} \\

        \midrule
        \midrule
        \multicolumn{10}{c}{\textit{Supervised Fine-Tuning}} \\
        \midrule
        \midrule

        $\pi_0^*$~\cite{black2410pi0} & Continuous &
        79.6 & 21.1 & 72.5 & 84.7 & 86.2 & 68.3 & 69.4 & 67.4 \\
        $\pi_{0.5}^*$~\cite{intelligence2025pi_} & Continuous &
        70.3 & \underline{41.7} & 81.1 & \textbf{97.3} & \textbf{94.6} & 71.8 & \textbf{84.9} & 75.7 \\
        OpenVLA-OFT+~\cite{fei2025libero} & Continuous &
        \underline{92.8} & 30.3 & \textbf{85.8} & 94.9 & 93.9 & \underline{89.3} & \underline{77.6} & \underline{79.6} \\

        \midrule
        \rowcolor{gray!20}
        \textbf{Ours} & \textbf{Hybrid} &
        \textbf{94.5} &
        \textbf{41.8} &
        \underline{83.2} &
        \underline{95.3} &
        \underline{94.3} &
        \textbf{93.7} &
        75.3 &
        \textbf{82.3} \\
        \bottomrule
    \end{tabular}
    }
    \caption{Results on the LIBERO-Plus benchmark under two settings: \textit{Zero-Shot Transfer}, where models trained on LIBERO are tested on LIBERO-Plus without fine-tuning, and \textit{Supervised Fine-Tuning}, where models are trained on the LIBERO-Plus training set. An asterisk (*) denotes results reproduced by us. The best results are highlighted in \textbf{bold}, and the second-best results are \underline{underlined}.}
    \vspace{-3mm}
    \label{tab:libero_plus}
\end{table*}

\noindent\textbf{LIBERO-Plus Benchmark.}
LIBERO-Plus introduces controlled perturbations across seven distinct dimensions, including variations in camera viewpoints, lighting conditions, background textures, object layouts, and robot initial states. We evaluate under both \textit{Zero-Shot Transfer}, where models trained on LIBERO are directly tested on LIBERO-Plus, and \textit{Supervised Fine-Tuning} on the LIBERO-Plus training set. We report the success rate across each of the seven perturbation dimensions and the overall average.

As detailed in Table~\ref{tab:libero_plus}, \modelname achieves state-of-the-art performance under both settings. Under Zero-Shot Transfer, our model attains 79.5\% average success rate, demonstrating strong robustness against diverse perturbations without explicit adaptation. Under Supervised Fine-Tuning, \modelname further improves to 82.3\%, surpassing baselines. Regarding resilience to visual domain shifts, results validate that the \sysone maintains robust intent guidance, ensuring the \systwo generates effective precise actions. Furthermore, for robot state initialization errors, the stable macro-directional guidance provided by \sysone enables the model to dynamically adjust action directions, demonstrating error recovery capabilities.

\subsection{Ablation Studies}
\label{sec:ablation}

In this section, we conduct extensive ablation studies on the LIBERO benchmark to provide a comprehensive analysis of \modelname. Specifically, our investigations focus on the following topics:

\begin{enumerate}
    \item \textbf{Architectural Effectiveness:} Dissect the contribution of individual model components.
    \item \textbf{Intent Granularity:} Analyze the impact of action intent granularity on model performance.
    \item \textbf{Training Strategy:} Validate the effectiveness of the dynamic curriculum training strategy.
    \item \textbf{Asynchronous Execution:} Investigate the influence of the \textit{Horizon Expansion Factor} ($M$).
\end{enumerate}

\noindent\textbf{Architectural Effectiveness.} To validate the effectiveness of individual model components, we evaluate three variants: Libra-Base (a standard monolithic VLA), Libra-VE (the baseline augmented with an auxiliary visual encoder), and Libra-Refinement (adopting the Coarse-to-Fine generation paradigm but without the auxiliary visual encoder). Detailed architectural specifications are provided in Appendix~\ref{appendix:model_implementation}.

\begin{table}[t]
    \centering
    \renewcommand{\arraystretch}{1.05}
    \setlength{\tabcolsep}{5pt}
    \resizebox{\columnwidth}{!}{
        \begin{tabular}{c|cc|cccc|c}
        \toprule
        Model & VE & Refine & Spatial & Object & Goal & Long & Avg. \\
        \midrule
        Libra-Base & \ding{55} & \ding{55} & 
        95.8 & 95.4 & 86.0 & 76.0 & 88.3 \\
        
        Libra-VE & \ding{51} & \ding{55} & 
        94.8 & 94.6 & 69.2 & 89.4 & 87.0 \\
        
        Libra-Refinement & \ding{55} & \ding{51} & 
        \textbf{98.6} & 98.6 & 96.0 & 87.0 & 95.1 \\
        \midrule
        
        \textbf{Full} & \ding{51} & \ding{51} & 
        \textbf{98.6} & \textbf{99.4} & \textbf{98.0} & \textbf{92.8} & \textbf{97.2} \\
        \bottomrule
        \end{tabular}
    }
    \caption{Ablation study on model components.}
    \vspace{-6mm}
    \label{tab:ablation_components}
\end{table}

The results in Table~\ref{tab:ablation_components} reveal three key insights into the architectural rationale of \modelname.

Comparing Libra-VE against the baseline Libra-Base, we observe that simply appending a visual encoder does not guarantee improvement. In fact, the average success rate declines to 87.0\%. Notably, performance on the Goal task plummets from 86.0\% to 69.2\%. This suggests that without structural constraints, the incorporation of dense visual features induces the model to learn visual shortcuts.

In contrast, while maintaining the same trainable parameters as Libra-Base, Libra-Refinement achieves a substantial leap in performance to 95.1\%. This validates that the core advantage stems from the Coarse-to-Fine generation paradigm, which lowers the overall learning difficulty, enabling the model to synthesize robot actions more effectively.

Moreover, full model surpasses Libra-Refinement, particularly in long-horizon tasks, by resolving the feature coupling bottleneck where the single VLM backbone struggles to balance the competing demands of high-level semantic reasoning and low-level geometric feature extraction. The introduction of an independent Visual Encoder achieves \textit{structural decoupling}: it specifically extracts geometric features for the refinement phase, effectively offloading the VLM. This allows the VLM to focus solely on semantic planning while the encoder handles precise actuation, resulting in a synergistic boost in robustness.

\noindent\textbf{Intent Granularity.} A critical hyperparameter in our architecture is the number of quantization bins ($N$) for the coarse action, which defines the granularity of the intent. We posit that an appropriate $N$ exists. If $N$ is too small, it fails to provide informative guidance, while an excessively large $N$ turns the coarse-level planning task into an intractable fine-grained classification problem, which fundamentally undermines our core motivation of decomposing the learning complexity via a coarse-to-fine hierarchical action generation strategy.
To verify the hypothesis, we conduct experiments across two model configurations: the Libra-Refinement variant and our full model.

\begin{table}[t]
    \centering
    \renewcommand{\arraystretch}{1.0}
    \setlength{\tabcolsep}{4.5pt}
    \resizebox{\columnwidth}{!}{
        \begin{tabular}{cc|c|cccc|c}
        \toprule
        VE & Refine & Bin ($N$) & Spatial & Object & Goal & Long & Avg. \\
        \midrule
        
        % Group 1: Refinement Only (No VE)
        \multirow{4}{*}{\ding{55}} & \multirow{4}{*}{\ding{51}} 
          &  2   & 95.8 & 97.6 & 95.4 & 80.2 & 92.3 \\
          & & 10  & \textbf{98.6} & \textbf{98.6} & \textbf{96.0} & \textbf{87.0} & \textbf{95.1} \\
          & & 50  & 96.0 & 96.4 & 78.2 & 79.4 & 87.5 \\
          & & 100 & 94.0 & 95.0 & 70.8 & 75.6 & 83.9 \\
        
        \midrule
        
        % Group 2: Full System (With VE)
        \multirow{4}{*}{\ding{51}} & \multirow{4}{*}{\ding{51}} 
          & 2   & 97.0 & 98.6 & 38.6 & 81.8 & 79.0 \\
          & & 10  & \textbf{98.6} & \textbf{99.4} & \textbf{98.0} & \textbf{92.8} & \textbf{97.2} \\
          & & 50  & 96.8 & 98.4 & 95.0 & 89.2 & 94.9 \\
          & & 100 & 95.4 & 96.8 & 92.8 & 90.4 & 93.9 \\
          
        \bottomrule
        \end{tabular}
    }
    \caption{Ablation study on coarse bin sizes ($N$).}
    \vspace{-6mm}
    \label{tab:ablation_bins}
\end{table}

\begin{figure}[t]
    \centering
    \includegraphics[width=\columnwidth]{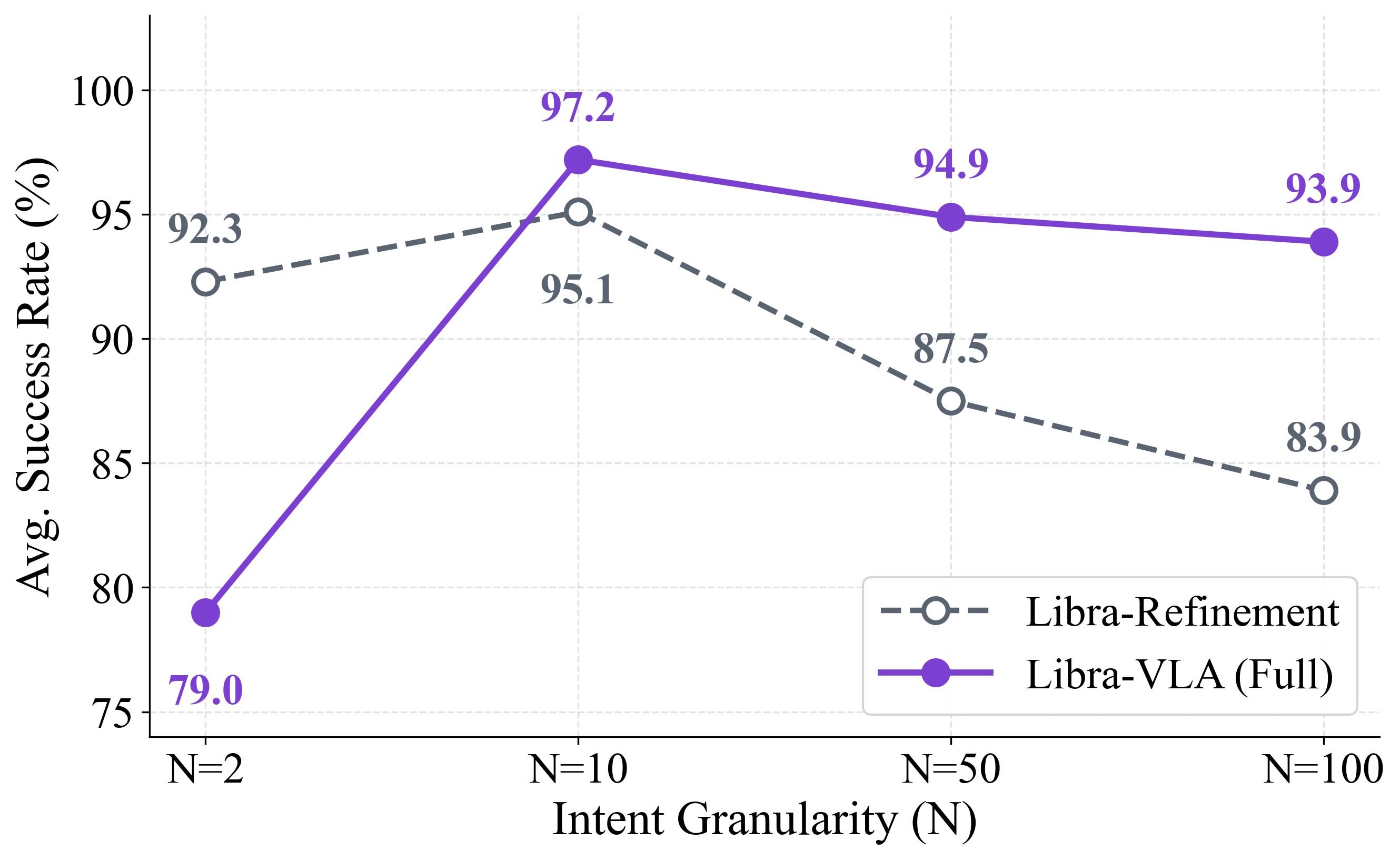}
    \caption{Performance trend with respect to the coarse bin size $N$.}
    \vspace{-6mm}
    \label{fig:inverted_u}
\end{figure}

As shown in Table~\ref{tab:ablation_bins}, we observe a consistent inverted-U performance trend in both settings, as visualized in Fig.~\ref{fig:inverted_u}.
This phenomenon can be fundamentally interpreted through the principle of \textit{complexity decomposition}. The bin size $N$ acts as a lever that shifts the distribution of learning difficulty between the two subsystems.

At extremely low granularity ($N=2$), the coarse action tokens suffer from insufficient information density. Due to the lack of informative guidance, the system effectively degenerates into a pure diffusion paradigm, forcing the \systwo to shoulder the entire burden of trajectory synthesis. The \sysone fails to provide a meaningful geometric anchor, leaving the refiner to solve the complex manipulation task almost single-handedly, resulting in suboptimal performance.

Conversely, employing excessive bin sizes ($N \ge 50$) shifts the paradigm towards high-precision discrete autoregression. This configuration drastically increases the number of discrete categories, disproportionately overwhelming the \sysone with the complexity of precise metric prediction. The resulting degradation in coarse prediction accuracy feeds erroneous directional guidance to the subsequent \systwo, triggering a cascading error that ultimately compromises the overall task success rate.

The ``Libra point'' is thus found at \textbf{$N=10$}, where the model achieves peak performance. This setting represents a learning equilibrium. The coarse tokens provide sufficiently informative guidance to effectively alleviate the learning load on \systwo, yet remain abstract enough to avoid imposing an excessive learning burden on the planner. By effectively distributing the workload, assigning broad intent modeling to the planner and local refinement to the refiner, this configuration maximizes the efficacy of the hierarchical architecture. A detailed analysis of the resulting training dynamics and convergence behavior is further provided in Section~\ref{sec:convergence}.

\noindent\textbf{Training Strategy.}
As described in Section~\ref{sec:systeLibra-VE}, we employ a dynamic curriculum strategy to stabilize training and enhance the robustness of the \systwo. To validate its effectiveness, we compare against two baselines: (1) \textit{Pure Teacher Forcing}, which always conditions on ground-truth coarse actions, and (2) \textit{No Teacher Forcing}, which always conditions on predicted coarse actions from \sysone.

\begin{table}[t]
    \centering
    \renewcommand{\arraystretch}{1.05}
    \setlength{\tabcolsep}{4pt}
    \resizebox{\columnwidth}{!}{
        \begin{tabular}{l|cccc|c}
        \toprule
        Training Strategy & Spatial & Object & Goal & Long & Avg. \\
        \midrule
        Pure Teacher Forcing & 96.6 & 99.2 & 95.8 & 92.4 & 96.0 \\
        No Teacher Forcing & 97.4 & 99.0 & 95.0 & 90.4 & 95.5 \\
        \midrule
        \textbf{Dynamic Curriculum (Ours)} & \textbf{98.6} & \textbf{99.4} & \textbf{98.0} & \textbf{92.8} & \textbf{97.2} \\
        \bottomrule
        \end{tabular}
    }
    \captionsetup{skip=5pt}
    \caption{Ablation study on the training strategy.}
    \vspace{-4mm}
    \label{tab:ablation_training}
\end{table}

As shown in Table~\ref{tab:ablation_training}, our dynamic curriculum strategy achieves the highest average success rate. Pure Teacher Forcing, while providing stable training gradients, suffers from a significant training-inference gap: the \systwo becomes over-reliant on perfect coarse inputs and fails to develop error-correction capabilities, leading to degraded performance when conditioned on imperfect predictions during inference. Conversely, No Teacher Forcing yields the lowest performance, particularly on long-horizon tasks, as the early-stage planning noise from the unconverged \sysone severely destabilizes the \systwo optimization. Our strategy effectively balances these two extremes: initial ground-truth forcing ensures stable early convergence, while the subsequent transition to predicted anchors exposes the \systwo to realistic planning noise, fostering robustness and consistency between training and inference distributions.

\begin{table}[t]
    \centering
    \renewcommand{\arraystretch}{1.1}
    \setlength{\tabcolsep}{3.5pt} 
    \resizebox{\columnwidth}{!}{
        \begin{tabular}{c|cccc|c|cc}
        \toprule
        Factor ($M$) & Spatial & Object & Goal & Long & Avg. & Latency (ms)  & Reduction  \\
        \midrule
        2 & \textbf{98.6} & 99.4 & \textbf{98.0} & 92.8 & \textbf{97.2} & 122 & 44.5\% \\
        3 & 97.6 & 99.2 & 94.2 & \textbf{93.8} & 96.2 & 112 & 49.1\% \\
        4 & 97.4 & \textbf{99.8} & 93.2 & \textbf{93.8} & 96.1 & 107 & 51.4\% \\
        5 & 97.8 & 98.8 & 92.0 & 92.4 & 95.3 & 104 & 52.7\% \\
        \bottomrule
        \end{tabular}
    }
    \caption{Ablation study on the \textit{Horizon Expansion Factor} ($M$).}
    \vspace{-6mm}
    \label{tab:ablation_horizon}
\end{table}

\noindent\textbf{Asynchronous Execution.}
We further investigate the impact of the \textit{Horizon Expansion Factor} ($M$), which governs the frequency of the asynchronous semantic planning updates. The results are summarized in Table~\ref{tab:ablation_horizon}. Serving as the baseline, Libra-Base records an inference time of 220 ms on Nvidia RTX 4090. A detailed comparison of the generation paradigms is provided in Appendix~\ref{appendix:model_implementation} to ensure fair evaluation.

As the expansion factor $M$ increases from 2 to 5, we observe a slight downward trend in the average success rate, declining from 97.2\% to 95.3\%. However, this performance degradation is not drastic. Even at $M=5$, the model maintains a high success rate of over 95\%. We attribute the robustness against higher expansion factors ($M$) to the spatial tolerance of our coarse quantization ($N=10$), which effectively accommodates the accumulated trajectory errors and state drift inherent in prolonged open-loop execution.
Increasing $M$ substantially reduces the average inference latency by amortizing the heavy computational cost of the VLM-based planning over more execution steps.

\subsection{Training Convergence Analysis}
\label{sec:convergence}

To further examine the convergence behavior of our coarse-to-fine design, we compare the intermediate rollout performance of Libra-VLA against the monolithic baseline Libra-Base at 10{,}000 training steps, i.e., one-third of the total training budget. Both models are trained on the same LIBERO data with the identical batch size and optimizer configuration described in Appendix~\ref{appendix:training_details}. We focus on intermediate rollout success rates to directly reflect task-level behavior.

\begin{table}[t]
    \centering
    \renewcommand{\arraystretch}{1.05}
    \setlength{\tabcolsep}{5pt}
    \resizebox{\columnwidth}{!}{
        \begin{tabular}{l|cccc|c}
        \toprule
        Method & Spatial & Object & Goal & Long & Avg. \\
        \midrule
        Libra-Base & 81.0 & 86.8 & 66.0 & 54.4 & 72.1 \\
        \textbf{Libra-VLA (Ours)} & \textbf{97.2} & \textbf{99.2} & \textbf{75.0} & \textbf{82.2} & \textbf{88.4} \\
        \bottomrule
        \end{tabular}
    }
    \caption{Success rates (\%) at 10,000 training steps on the LIBERO benchmark. Results are averaged over 500 independent rollouts per task suite, following the same evaluation protocol as Appendix~\ref{appendix:hybridvla}.}
    \vspace{-4mm}
    \label{tab:convergence_10k}
\end{table}

As shown in Table~\ref{tab:convergence_10k}, at 10k steps Libra-VLA reaches an average success rate of 88.4\%, exceeding Libra-Base by 16.3 points under identical training conditions. The gap is most pronounced on the Long suite, where the longer-horizon planning is more sensitive to the optimization load of grounding high-level semantics to continuous actions.

\begin{figure}[t]
    \centering
    \includegraphics[width=\columnwidth]{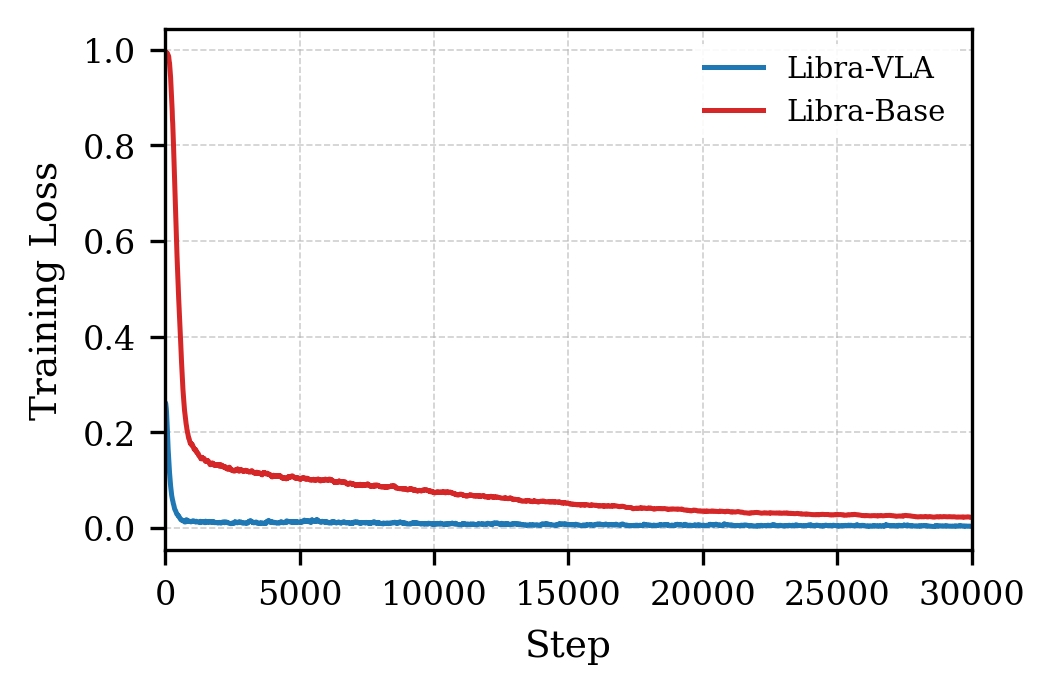}
    \caption{Training loss curves of Libra-VLA and Libra-Base on LIBERO.}
    \vspace{-6mm}
    \label{fig:loss_curve}
\end{figure}

\begin{figure}[t]
    \centering
    \includegraphics[width=\columnwidth]{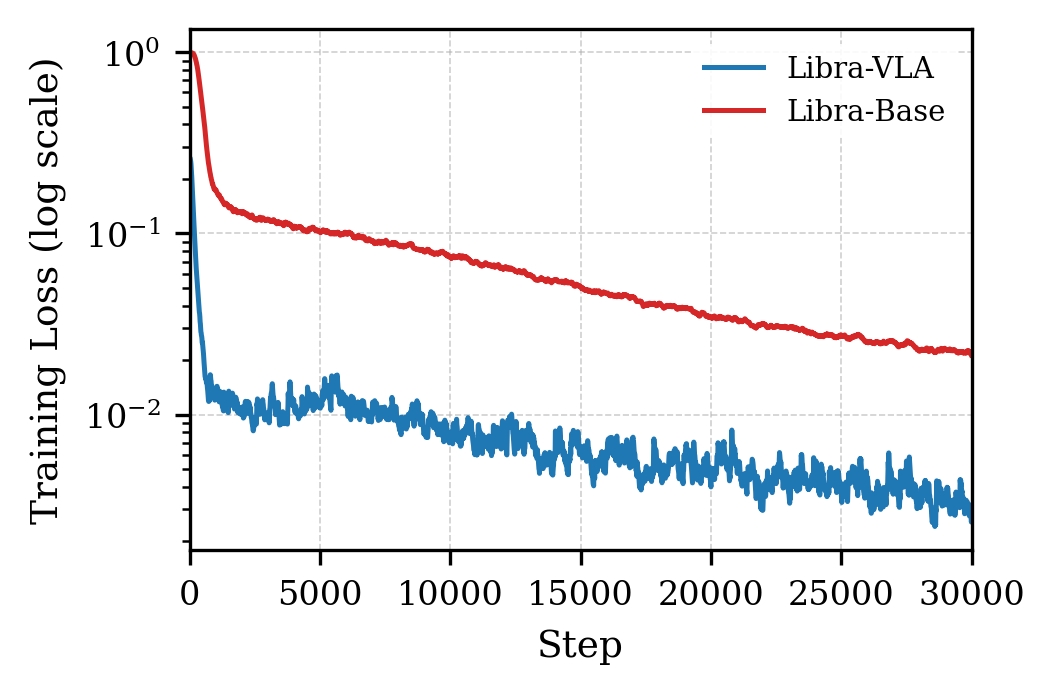}
    \caption{Training loss curves plotted on a logarithmic scale.}
    \vspace{-6mm}
    \label{fig:log_loss_curve}
\end{figure}

We attribute this intermediate gap to the workload decoupling inherent in our design. The \sysone focuses on low-frequency macro-intent prediction, while the \systwo performs continuous micro-alignment conditioned on the predicted coarse anchors. These anchors narrow the effective search space of the \systwo, which alleviates the optimization burden observed in monolithic generation. This workload decoupling is directly reflected in the training loss curves shown in Figs.~\ref{fig:loss_curve} and~\ref{fig:log_loss_curve}, which plot the MSE loss on the continuous action outputs, corresponding to the fine-action generation in Libra-VLA and the monolithic action generation in Libra-Base. The log-scale view more clearly reveals the low-loss regime where the linear-scale view in Fig.~\ref{fig:loss_curve} saturates visually. At 10k training steps, this continuous-action MSE loss of Libra-VLA decreases to approximately $0.01$, while that of Libra-Base remains noticeably higher at around $0.07$, consistent with the workload-balance argument discussed in Section~\ref{sec:ablation}.

We further note a short-lived rise in the Libra-VLA curve around step $5{,}000$. This corresponds to the transition point of our dynamic curriculum strategy (Section~\ref{sec:ablation}): once the coarse-action prediction accuracy of the \sysone surpasses a preset threshold, the input to the \systwo is switched from ground-truth coarse actions to those predicted by the \sysone, which briefly exposes the refiner to imperfect anchors. The loss soon resumes its downward trend within a limited number of steps, indicating that the \systwo gradually learns to compensate for noisy coarse inputs, which is in line with the error-correction behavior discussed in Section~\ref{sec:ablation}.

\subsection{Real-World Experiments}
\label{sec:real_world_exp}

\begin{figure}[t]
    \centering
    \includegraphics[width=\columnwidth]{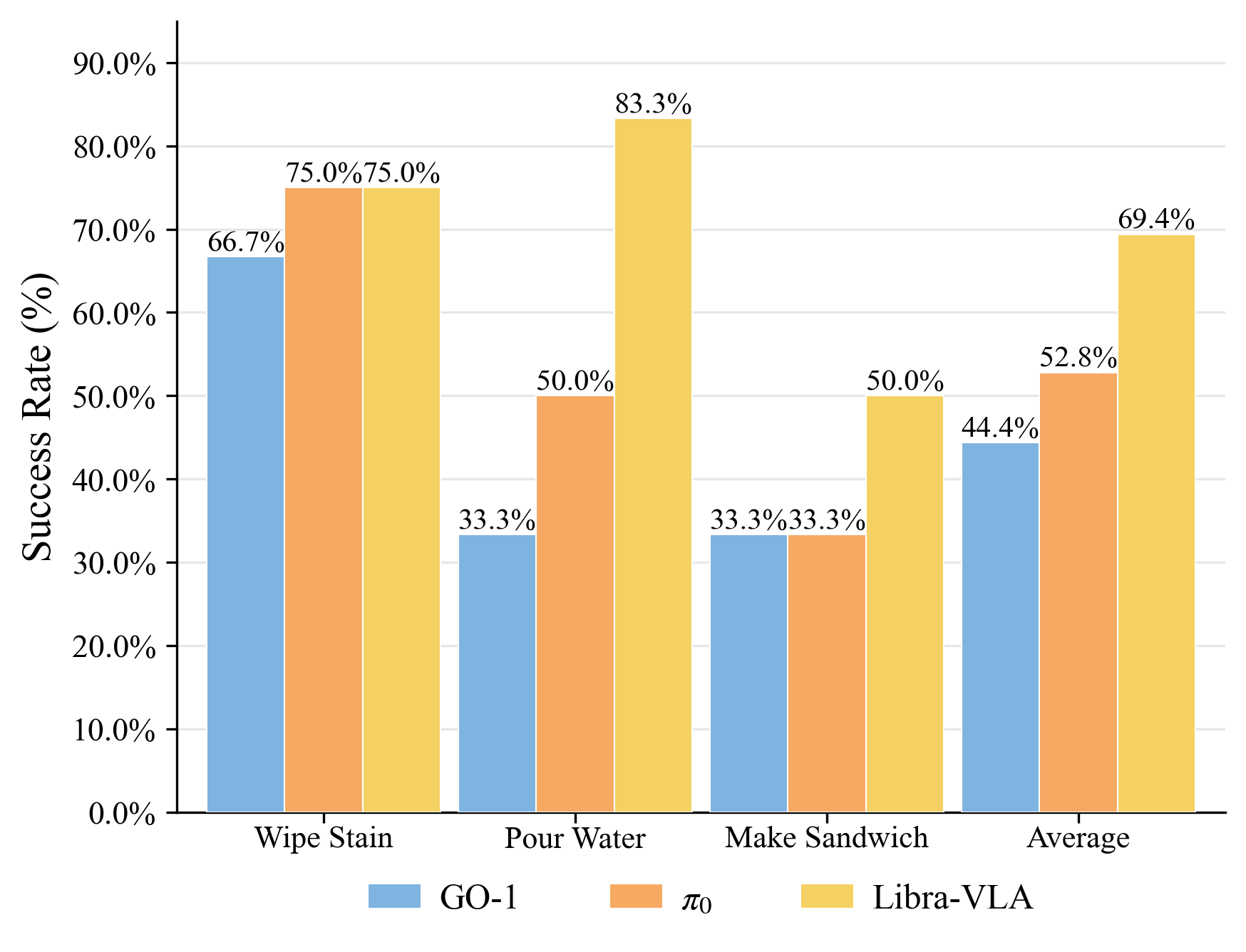}
    \caption{Results of real-world experiments.}
    \vspace{-6mm}
    \label{fig:real}
\end{figure}

We conducted several real-world experiments to further evaluate the effectiveness and robustness of our proposed Libra-VLA in unstructured physical environments. We conducted three long-horizon tasks, e.g., ``Wipe Stain'', ``Pour Water'', and ``Make Sandwich'', which necessitate sustained temporal coherence and high-precision spatial grounding. 
Detailed execution protocols for these tasks are as follows. ``Wipe Stain'' requires the robot to first identify and grasp a sponge randomly positioned on the table, then clean the target stain randomly scattered. ``Pour Water'' task requires the robot to grasp a kettle positioned on the table, localize the target cup, pour an appropriate amount of water, and subsequently return the kettle to its designated coaster. ``Make Sandwich'' is a complex, multi-stage manipulation task comprising four distinct sub-tasks. This task entails sequentially stacking ingredients (e.g., bread, meat, lettuce) onto a plate to assemble a complete sandwich. More details about real-world tasks can be found in Appendix~\ref{appendix:real_world_tasks}.

To comprehensively assess performance, we compare our method against competitive baselines \textbf{$\pi_0$} and Go-1, reporting both the \textit{Single-task Success Rate} and the \textit{Average Success Rate}. The result is shown in Fig.~\ref{fig:real}.

\section{Conclusion}
\label{sec:conclusion}

In this paper, we presented \modelname, which pioneers a novel \textit{Coarse-to-Fine generation paradigm} for action generation. Grounded in a Hybrid Action Space, this framework decomposes manipulation into discrete macro-intents and continuous micro-residuals. Through extensive analysis, we identified the ``Libra Point'', a balanced granularity equilibrium that effectively balances the learning complexity between two generation phases. Empirical results demonstrate that this decoupled architecture not only achieves state-of-the-art success rates on complex benchmarks but also significantly reduces inference latency for real-time control.

\section*{Limitations}

\label{sec:limitation}
While Libra-VLA demonstrates strong robustness in asynchronous manipulation, we observe a marginal attenuation in performance as the frequency of asynchronous execution increases. Although this performance decline is slight rather than drastic, it indicates a limitation of the current asynchronous strategy. Our current asynchronous strategy adopts a relatively straightforward protocol where all predicted coarse macro-intents are utilized sequentially. This approach lacks a dynamic verification mechanism to filter out potentially suboptimal anchors during long-horizon execution. To address this, our future work aims to integrate a real-time confidence estimation mechanism. This will allow the system to dynamically assess the reliability of the current macro-intent and trigger a regeneration of the coarse action if the confidence falls below a critical threshold, thereby further enhancing adaptability in complex scenarios.

\section*{Ethical Considerations}

We acknowledge several ethical considerations and potential risks associated with our research.

\noindent\textbf{Physical Safety.} The primary risk in deploying VLA models lies in the potential for unpredictable physical actions, which could lead to hardware damage or safety hazards in unstructured environments. To mitigate this, all real-world experiments in this study were conducted in a controlled laboratory setting under strict human supervision, with immediate emergency stop mechanisms in place. We emphasize that future deployment of such models in open-ended environments requires rigorous safety testing and fail-safe protocols.

\noindent\textbf{Data Privacy.} We strictly adhere to ethical data usage standards. While the dataset encompasses real-world scenarios, rigorous filtering protocols were applied to protect privacy. We ensured that personally identifiable information (PII), particularly human faces, has been anonymized or excluded from the training and evaluation sets. No offensive content is present in the data.

\noindent\textbf{Model Bias.} As our model builds upon pre-trained Vision-Language Models (VLMs), it may inherit biases present in the large-scale pre-training data. While we focus on manipulation tasks, users should be aware of these potential biases when interpreting the model's high-level reasoning capabilities.

\bibliography{custom}

\appendix

\section{Training Data}
\label{appendix:training_data}

In this section, we present a comprehensive introduction of the simulation benchmarks and real-world tasks in this work.
\subsection{LIBERO Benchmark Suite}
 LIBERO is a large-scale, procedurally generated benchmark designed to assess knowledge transfer and lifelong learning capabilities in robot manipulation. It provides a diverse set of tasks that require the agent to master both declarative knowledge and procedural knowledge.

The benchmark consists of four distinct task suites, each curating specific distribution shifts to evaluate different facets of the model's generalization ability:

\begin{itemize}
    \item \textbf{LIBERO-Spatial}: This suite contains 10 tasks where the robot must perform the same manipulation primitive but operates under varying spatial layouts. The objects remain consistent, but their relative positions change, requiring the agent to possess robust spatial reasoning capabilities.
    
    \item \textbf{LIBERO-Object}: Comprising 10 tasks, this suite fixes the spatial layout and task structure but introduces diverse object instances. The agent must generalize its manipulation skills across different visual textures and geometries, testing its object-centric visual grounding.
    
    \item \textbf{LIBERO-Goal}: This suite includes 10 tasks that share the same workspace and object arrangement but differ in the semantic goals. This evaluates the agent's ability to follow and execute distinct instructions within an identical visual context.
    
    \item \textbf{LIBERO-Long}: As the most challenging suite, it consists of 10 long-horizon tasks. Each task requires the sequential execution of multiple primitives to complete a complex objective. This suite rigorously tests the model's ability to handle temporal dependencies and multi-stage planning.
\end{itemize}

\begin{figure}[t]
    \centering
    \includegraphics[width=\columnwidth]{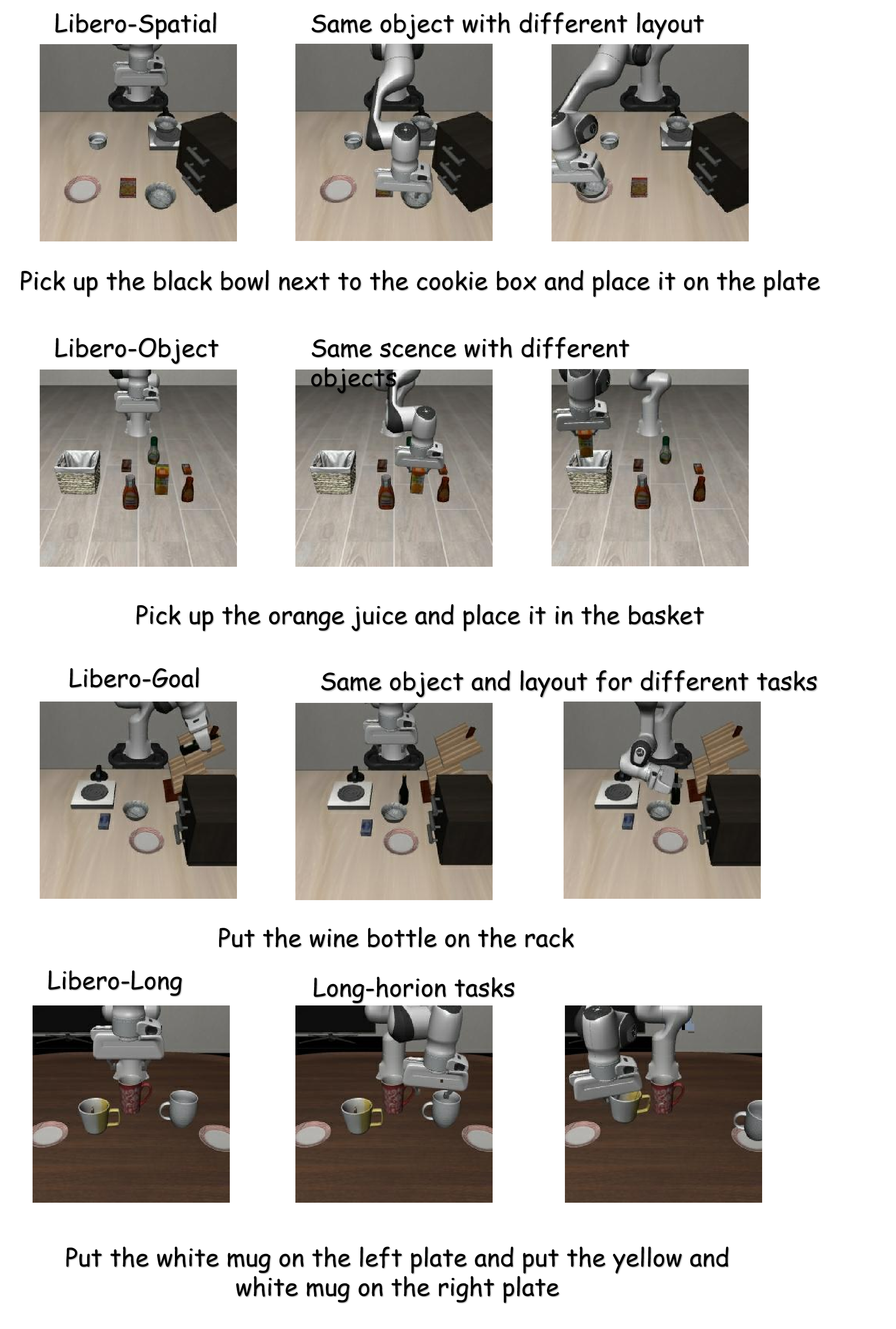}
    \caption{Examples of the LIBERO benchmark.}
    \vspace{-6mm}
    \label{fig:LIBERO}
\end{figure}

 For the training data, LIBERO dataset contains 1,693 episodes and 273,465 frames, recorded at a fixed 10 Hz. Our model is trained for 30k steps with global batch size 128. Examples of the LIBERO benchmark are shown in Fig.~\ref{fig:LIBERO}.

\subsection{LIBERO-Plus Benchmark Suite}
\label{appendix:libero_plus_setup}

\begin{figure*}[t]
    \centering
    \includegraphics[width=\textwidth]{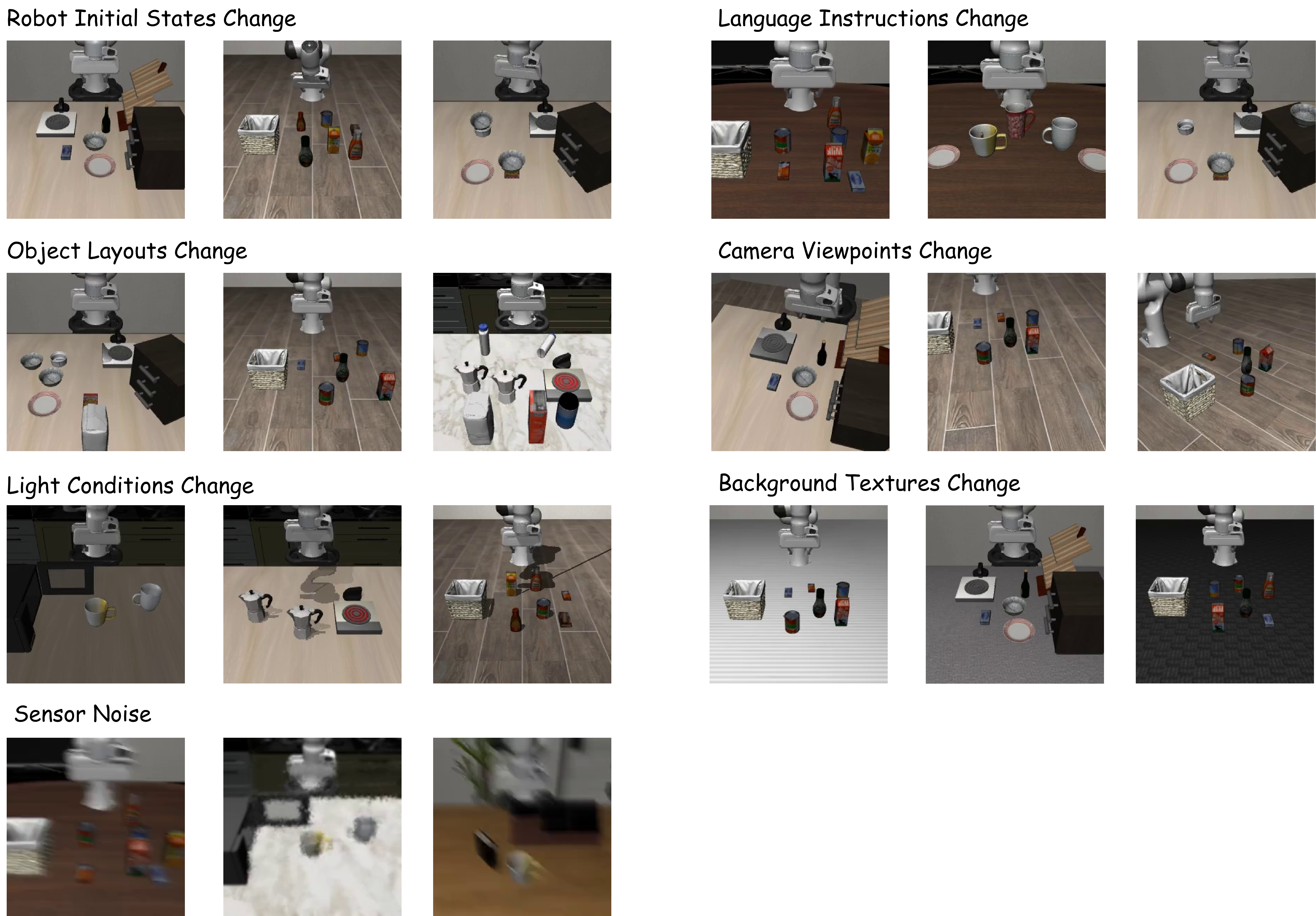}
    \caption{Examples of the LIBERO-Plus benchmark.}
    \vspace{-3mm}
    \label{fig:lbplus}
\end{figure*}

While the standard LIBERO benchmark evaluates the agent's ability to transfer knowledge across varying task semantics and spatial layouts, it operates within a relatively "clean" and idealized visual environment. To rigorously assess the robustness of our proposed model against real-world nuisance variables, we further employ the LIBERO-Plus benchmark.

LIBERO-Plus extends the evaluation protocol by performing a systematic vulnerability analysis across seven distinct perturbation dimensions. These perturbations are designed to mimic distribution shifts and uncertainties inherent in unstructured real-world deployments:

\begin{itemize}
    \item \textbf{Object Layouts}: This perturbation introduces variations in the initial spatial configuration of task-relevant objects, testing the model's ability to adapt to unseen object arrangements beyond the training distribution.
    
    \item \textbf{Camera Viewpoints}: This perturbation introduces random jitter and offsets to the camera's extrinsic parameters (position and orientation). This evaluates the model's tolerance to viewpoint shifts and its ability to maintain spatial consistency without overfitting to a fixed camera pose.
    
    \item \textbf{Robot Initial States}: The initial joint configuration or end-effector pose of the robot is randomized. This challenges the policy to recover from diverse starting conditions and successfully plan trajectories to the target.
    
    \item \textbf{Language Instructions}: To assess semantic robustness, the task instructions are rephrased using synonyms or different sentence structures while preserving the original semantic meaning (e.g., changing "pick up" to "grasp").
    
    \item \textbf{Light Conditions}: This perturbation simulates uncontrolled illumination environments by altering the intensity, direction, and color temperature of the scene's lighting, testing the model's invariance to photometric shifts.
    
    \item \textbf{Background Textures}: The visual appearance of the workspace (e.g., table surface, background walls) is randomized with diverse textures, evaluating the model's ability to perform figure-ground separation and ignore background distractors.
    
    \item \textbf{Sensor Noise}: Gaussian noise or other forms of signal corruption are injected into the visual observations or proprioceptive states, mimicking high-ISO camera noise or sensor degradation in physical hardware.
\end{itemize}

LIBERO-Plus dataset provides 14,347 episodes and 2,238,036 frames, captured at 20 Hz. Our model is trained for 50k steps with global batch size 128. Examples of the LIBERO-Plus benchmark are shown in Fig.~\ref{fig:lbplus}.
\subsection{Real-World Tasks}
\label{appendix:real_world_tasks}

\begin{figure}[t]
    \centering
    \includegraphics[width=\columnwidth]{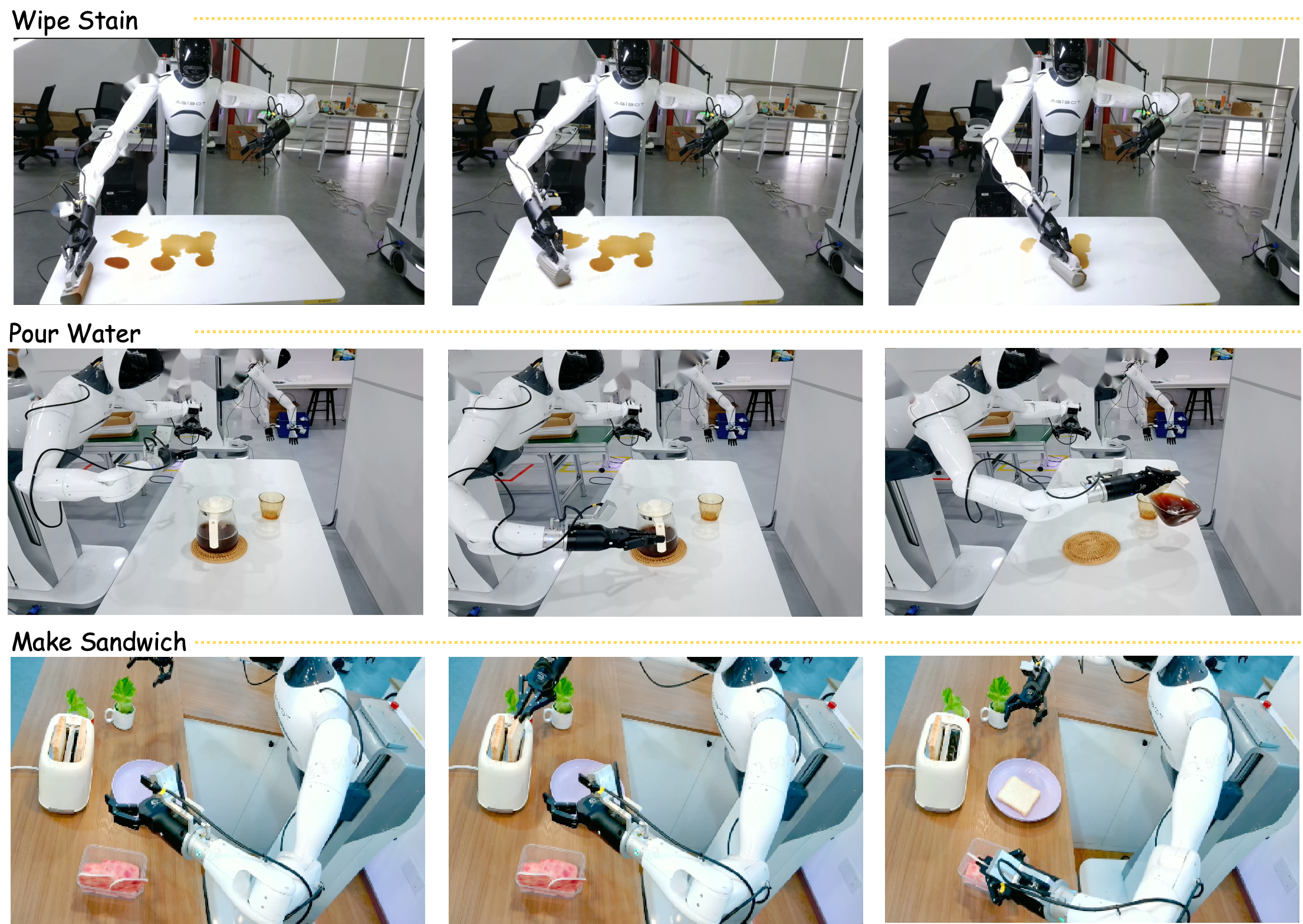}
    \caption{Visualization of real-world tasks.}
    \vspace{-6mm}
    \label{fig:real_vis}
\end{figure}

All real-world data collection and evaluation experiments were conducted on AgiBot G1 robot platform. Real-world tasks evaluated in our experiments are shown in Fig.~\ref{fig:real_vis}.

In the ``Wipe Stain'' task, the robot is required to first visually localize and grasp a sponge placed on the tabletop. Following the grasp, the agent must identify the specific location of the stain and manipulate the sponge to perform the erasing action. The ``Wipe Stain'' task consists of $177$ episodes totaling $356,316$ frames.

For the ``Pour Water'' task, the robot is tasked with initially grasping a kettle, maneuvering it to an appropriate position above a target cup, dispensing an appropriate amount of water, and subsequently returning the kettle to its original location. This task comprises $1,821$ episodes with $5,062,506$ frames.

``Make Sandwich'' presents a long-horizon assembly challenge, which demands the robot to sequentially retrieve distinct ingredients—specifically bread, meat, and lettuce—and vertically stack them onto a serving plate to assemble a complete sandwich. The ``Make Sandwich'' dataset contains $452$ episodes comprising $1,222,087$ frames.

\section{Evaluation Metrics}
\label{sec:evaluation_metrics}

To comprehensively assess the performance of our method, we employ rigorous evaluation protocols. For real-world comparisons in particular, all compared models are evaluated under identical physical setups with pre-marked target positional configurations on the workspace. All rollouts for different models are further conducted consecutively within a short time window by the same human testers, robot, and props, so that lighting and other environmental conditions remain as consistent as possible across models.

\subsection{Simulation Benchmarks}
For the LIBERO and LIBERO-Plus simulation benchmarks, we strictly adhere to the official evaluation protocols to ensure a fair and consistent comparison with prior works. Specifically, we utilize the official testing scripts and maintain identical parameter settings as defined in the standard benchmark suite. This ensures that all reported results reflect the model's true capability under standardized conditions.

 Furthermore, to rigorously evaluate real-time performance of \modelname, we analyze the average inference latency across different Horizon Expansion Factor $M$. We define the average latency as the total duration of a complete asynchronous execution cycle divided by the number of execution steps within that cycle.

\subsection{Real-World Experiments}
For real-world evaluations, we define specific experimental setups and success criteria for each task as follows.

\noindent\textbf{Wipe Stain.} 
To test the model's generalization to spatial variations, the sponge is initialized at three distinct initial positions, while the target stain is randomly located at one of four different positions on the table. A single trial is considered successful if the robot successfully grasps the sponge and completely wipes the target stain.

\noindent\textbf{Pour Water.} 
In this task, we vary the initial position of the kettle across two locations and the target cup across three different locations. A trial is considered successful if the robot securely grasps the kettle and successfully pours an adequate amount of water into the cup.

\noindent\textbf{Make Sandwich.} 
This is a long-horizon task where all ingredients are placed at fixed initial positions. The robot acts as a chef and is required to sequentially grasp and place four ingredients onto a plate to assemble a complete sandwich. The entire task is considered successful only if all four sequential sub-tasks are correctly executed. Given the complexity of this long-horizon process, we allow a maximum of one retry for a specific sub-task if a failure occurs during execution.

\section{Implementation Details}
\label{appendix:model_implementation}
\begin{table*}[t]
    \centering
    \renewcommand{\arraystretch}{1.2}
    \setlength{\tabcolsep}{4pt} 
    \resizebox{\textwidth}{!}{
        \begin{tabular}{l|c|c|c|c|c}
        \toprule
        \textbf{Model} & \textbf{VLM Backbone} & \textbf{Action Expert Layers} & \textbf{Extra VE} & \textbf{Trainable Params} & \textbf{Total Params} \\
        \midrule
        Libra-Base & InternVL2.5-2B & $N=24$ & -- & 2287M & 2591M \\
        Libra-VE & InternVL2.5-2B & $N=24$ & SigLIP & 2738M & 3042M \\
        Libra-Refinement & InternVL2.5-2B & $N=12~/~N=12$ & -- & 2287M & 2591M \\
        Libra-VLA (Full) & InternVL2.5-2B & $N=12~/~N=12$ & SigLIP & 2738M & 3042M \\
        \bottomrule
        \end{tabular}
    }
    \caption{Detailed architectural specifications and parameter statistics.}
    \vspace{-4mm}
    \label{tab:model_specs}
\end{table*}

In this section, we provide a more comprehensive description of the implementation details for our proposed Libra-VLA architecture, alongside the specific configurations of the variant models employed in our comparative analysis.

We instantiate the VLM backbone of Libra-VLA with InternVL2.5-2B. Structurally, both the Parallel Coarse-Action Head and the Fine-Action Head are implemented as transformer blocks comprising $N=12$ attention layers, the hidden state dimension of which is set to $1024$, corresponding to half of the VLM backbone's hidden size. We employ SigLIP as the visual encoder within the Action Refiner. Unless explicitly stated otherwise, all results on simulation benchmarks are obtained with a Horizon Expansion Factor of $M=2$ and an action chunk size of $H_{chunk}=5$, resulting in a macro-horizon of $L_{macro}=10$. For real-world evaluations, we maintain the Horizon Expansion Factor at $M=2$ while increasing the action chunk size to $H_{chunk}=20$, resulting in a macro-horizon of $L_{macro}=40$. 

Furthermore, we clarify the structural configurations of the ablation variants. Libra-Base adopts a standard VLA architecture like $\pi_0$, employing a monolithic Action Expert comprising 24 stacked attention layers attached to the VLM backbone. For Libra-Refinement, we maintain the fundamental model structure but allocate 12 attention layers each to the Parallel Coarse-Action Head and the Fine-Action Head. Consequently, this architecture essentially decouples the monolithic expert into two specialized modules, ensuring that Libra-Base and Libra-Refinement possess an identical number of trainable parameters. Libra-VE builds upon the Libra-Base baseline by incorporating an additional visual encoder (SigLIP) to provide auxiliary visual inputs, while retaining the monolithic 24-layer Action Expert configuration.

The Libra-Base, Libra-VE, and Libra-Refinement models all adopt a synchronous inference mode, where perceptual processing and action generation occur sequentially within a single control cycle. In contrast, the full Libra-VLA architecture employs an asynchronous execution strategy. The specific structural configurations and parameter statistics for Libra-VLA and its ablation variants are detailed in Table~\ref{tab:model_specs}.

We further clarify the generation paradigms of Libra-Base and Libra-VLA to facilitate fair inference latency comparison. Libra-Base adopts the same architecture as $\pi_0$, utilizing a monolithic diffusion-based action generation process rather than autoregressive decoding. In contrast, the generation process in Libra-VLA is divided into two stages: (1) \textit{Coarse Action Generation}, where the \sysone employs bidirectional parallel decoding to generate all coarse action tokens simultaneously in a single forward pass; and (2) \textit{Fine Action Generation}, where the \systwo generates fine-grained actions using diffusion, iteratively refining Gaussian noise into precise robot control commands conditioned on the coarse intents. The faster inference speed of Libra-VLA is primarily attributed to three factors: the asynchronous execution strategy amortizes the expensive VLM forward pass over multiple control steps; the \systwo contains fewer attention layers than the monolithic expert, reducing per-step computation; and the bidirectional parallel decoding in the \sysone avoids the sequential bottleneck of autoregressive generation.

\section{Training Details}

\label{appendix:training_details}

In this section, we provide a comprehensive description of the training implementation for the proposed framework. Unless explicitly stated otherwise, the training of all model variants adheres to the standardized protocols and hyperparameter settings described below.

For the simulation benchmarks, we train the model for $30$k steps on LIBERO and $50$k steps on LIBERO-Plus, employing a consistent global batch size of $128$. 

For real-world tasks, the training configurations are specified as follows: the ``Wipe Stain'' task is trained for $40$k steps with a global batch size of $96$, the ``Pour Water'' task for $30$k steps with a global batch size of $128$, and the ``Make Sandwich'' task for $50$k steps with a global batch size of $96$.

To ensure training stability and convergence, we utilize a peak learning rate of $2 \times 10^{-5}$. The learning rate is dynamically adjusted using a cosine-decay scheduler, accompanied by a linear warmup phase spanning the initial 1,000 steps. To regularize the model and mitigate overfitting, we apply a weight decay of $0.01$. For \modelname and its ablation variants, the vision encoder within the VLM remains frozen during training, while all other model components are trainable. No large-scale robot-data pretraining is used at any stage: the VLM backbone inherits InternVL2.5-2B weights, the System 1 visual encoder inherits standard SigLIP weights, and all action-related modules (the Parallel Coarse-Action Head and the Fine-Action Head) are trained from scratch on the target task data. For the reproduced $\pi_0$ and $\pi_{0.5}$ on LIBERO-Plus, we train the vision encoder and the action expert. Regarding the computational infrastructure, all models are trained on a cluster equipped with 8 NVIDIA H100 GPUs. Furthermore, we employ bfloat16 mixed-precision training to optimize memory efficiency and computational throughput without compromising numerical stability.

\section{Further Comparison}
\label{appendix:hybridvla}

To empirically validate the advantage of our hierarchical action generation over flat hybrid approaches, we conduct a direct quantitative comparison against HybridVLA on the LIBERO benchmark. We use the official open-source codebase of HybridVLA without any modification to its training or inference logic. To ensure fairness, both models are trained under identical configurations: the same LIBERO training data, an identical global batch size of 128, and a fixed training budget of 30,000 steps. We perform 500 independent rollout evaluations for each task suite.

\begin{table}[h]
    \centering
    \renewcommand{\arraystretch}{1.0}
    \setlength{\tabcolsep}{4.5pt}
    \resizebox{\columnwidth}{!}{
        \begin{tabular}{l|cccc|c}
        \toprule
        Method & Spatial & Object & Goal & Long & Avg. \\
        \midrule
        HybridVLA & 24.4 & 48.6 & 38.4 & 20.4 & 32.9 \\
        \textbf{Libra-VLA (Ours)} & \textbf{98.6} & \textbf{99.4} & \textbf{98.0} & \textbf{92.8} & \textbf{97.2} \\
        \bottomrule
        \end{tabular}
    }
    \captionsetup{skip=2pt}
    \caption{Comparison with HybridVLA on LIBERO.}
    \vspace{-5mm}
    \label{tab:hybridvla}
\end{table}

As shown in Table~\ref{tab:hybridvla}, under aligned training overheads, Libra-VLA comprehensively outperforms HybridVLA. Due to the training burden imposed by its flat architecture and high-precision hybrid generation, HybridVLA exhibits a significantly slower convergence rate, achieving an average success rate of only 32.9\% at 30k steps. In contrast, Libra-VLA achieves 97.2\% by decomposing the learning complexity via the coarse-to-fine hierarchy, demonstrating substantially faster convergence and higher performance.

\section{LLM Usage Statement}

In this paper, Large Language Models (LLMs) were used exclusively for linguistic polishing and grammatical correction to enhance readability. None of the technical methodology, implementation details, or experimental results were generated by LLM.

\end{document}